\documentclass[letterpaper, 10 pt, conference]{ieeeconf}
\IEEEoverridecommandlockouts
\usepackage{amsmath,amsfonts}
\usepackage{algorithmic}
\usepackage{algorithm}
\usepackage{array}
\usepackage[caption=false,font=normalsize,labelfont=sf,textfont=sf]{subfig}
\usepackage{textcomp}
\usepackage{stfloats}
\usepackage{url}
\usepackage{verbatim}
\usepackage{cite}
\usepackage{graphicx}
\graphicspath{{figures/}}

\usepackage{booktabs}

\usepackage{upgreek}
\usepackage{amssymb}
\usepackage[T1]{fontenc}
\usepackage{epstopdf}
\usepackage{amsmath, bm}
\usepackage[colorlinks,linkcolor=blue,anchorcolor=blue,citecolor=blue]{hyperref}
\usepackage{color,soul}
\usepackage{epsfig}
\usepackage{multirow}
\usepackage{mathptmx}
\usepackage{extarrows}
\usepackage{makecell}
\usepackage{ulem}
\usepackage{xcolor}
\usepackage{eso-pic}

\usepackage[authorsmultiline]{latexml}

\newcommand{\vect}[1]{\textbf{\textit{#1}}}

\definecolor{ysyblue}{rgb}{0.1, 0.3, 0.8}

\title{\LARGE \bf
Fluorescence-enhanced Whisker Array with Vision-based \\ Deformation Analysis for Underwater Source Localization}

\iflatexml
\author{%
\IEEEauthorblockN{Xiaochi Xie}%
\IEEEauthorblockA{Department of Mechanical Engineering, Stanford University, USA\\
\texttt{xxie15@stanford.edu}\\
Xiaochi Xie and Hao Li contribute equally.}%
\and
\IEEEauthorblockN{Hao Li}%
\IEEEauthorblockA{Department of Mechanical Engineering, Stanford University, USA\\
\texttt{li2053@stanford.edu}\\
Xiaochi Xie and Hao Li contribute equally.}%
\and
\IEEEauthorblockN{Shixuan Zhao}%
\IEEEauthorblockA{Department of Mechanical Engineering, Stanford University, USA\\
\texttt{sxuan@stanford.edu}}%
\and
\IEEEauthorblockN{Siyue Yao}%
\IEEEauthorblockA{Department of Mechanical Engineering, Stanford University, USA\\
State Key Laboratory of Mechanical Systems and Vibration; META Robotics Institute, Shanghai Jiao Tong University, Shanghai 200240, China\\
Corresponding: \texttt{sy.yao@sjtu.edu.cn}}%
\and
\IEEEauthorblockN{Shuran Song}%
\IEEEauthorblockA{Department of Electrical Engineering, Stanford University, USA\\
\texttt{shuran@stanford.edu}}%
\and
\IEEEauthorblockN{Mark R. Cutkosky}%
\IEEEauthorblockA{Department of Mechanical Engineering, Stanford University, USA\\
\texttt{cutkosky@stanford.edu}}%
}
\else
\author{Xiaochi Xie$^{1,\dagger}$, 
Hao Li$^{1,\dagger}$, 
Shixuan Zhao$^{1}$, 
Siyue Yao$^{1,3,4,*}$,
Shuran Song$^{2}$, 
and Mark R. Cutkosky$^{1}$
\thanks{$\dagger$ These authors contribute equally} 
\thanks{$^{1}$ Department of Mechanical Engineering, Stanford University, USA 
({\tt\small xxie15@stanford.edu}, {\tt\small li2053@stanford.edu}, {\tt\small sxuan@stanford.edu}, {\tt\small cutkosky@stanford.edu})}
\thanks{$^{2}$ Department of Electrical Engineering, Stanford University, USA 
({\tt\small shuran@stanford.edu})}
\thanks{$^{3,4}$ State Key Laboratory of Mechanical Systems and Vibration; META Robotics Institute, Shanghai Jiao Tong University, Shanghai 200240, China
(Corresponding: {\tt\small sy.yao@sjtu.edu.cn})}
}

\fi

\begin{document}
\AddToShipoutPictureBG*{\AtPageUpperLeft{\raisebox{-0.42in}{\makebox[\paperwidth][c]{\parbox{0.8\paperwidth}{\centering\footnotesize This work has been submitted to the IEEE for possible publication. Copyright may be transferred without notice, after which this version may no longer be accessible.}}}}}
\begin{sloppypar}

\maketitle
\thispagestyle{empty}
\pagestyle{empty}

\begin{abstract}
Deep-water biological observation is essential for understanding marine organisms and their interactions with the environment. 
However, conventional optical and acoustic approaches can introduce stimuli that alter animal behavior and bias biological observations. 
This paper proposes a fluorescence-enhanced whisker array sensing system that pinpoints underwater hydrodynamic sources through local optical readout rather than direct  source imaging.
Five spatially oriented whiskers, fabricated with nitinol cores and fluorescent urethane shells, are integrated with ultraviolet excitation and a monocular camera. 
Image enhancement and segmentation are applied to track the whisker deformation. 
A lightweight convolutional neural network captures temporal and cross-whisker features from $2 \, s$ sequences to estimate source localization. 
Pool experiments achieve a mean spatial localization error of $88 \, mm$, with $73.5 \, mm$ in radius and $2.5^{\circ}$ in angle, across a test region of $600 \, mm$ with $\pm 30^{\circ}$. 
Real-time localization of a moving thruster demonstrates the capability of the proposed method in dynamic scenarios, highlighting its potential for integration into underwater robots for hydrodynamic source detection, localization, and tracking in low-light environments. 
\end{abstract}

\section{Introduction}\label{Sec::Intro}

Deep-water biological and environmental observation is essential for understanding marine organisms and their interactions with the surrounding environment~\cite{yoerger2021mesobot}. 
However, conventional underwater observation,
commonly relies on vision-based sensing with artificial illumination~\cite{widder2005redlight} or active acoustic sensing such as sonar and scientific echosounders~\cite{cholewiak2017echosounders}, may alter creature behavior and consequently bias biological observations~\cite{ludvigsen2018light, berge2020light}. 
Such disturbances become increasingly problematic in deep water~\cite{benoitbird2023observer}. 
These findings motivate sensing approaches that acquire sufficient information under low ambient illumination while minimizing observation-induced disturbance. 

\begin{figure}[!t]
	\centering
	\includegraphics[width=0.99\columnwidth]{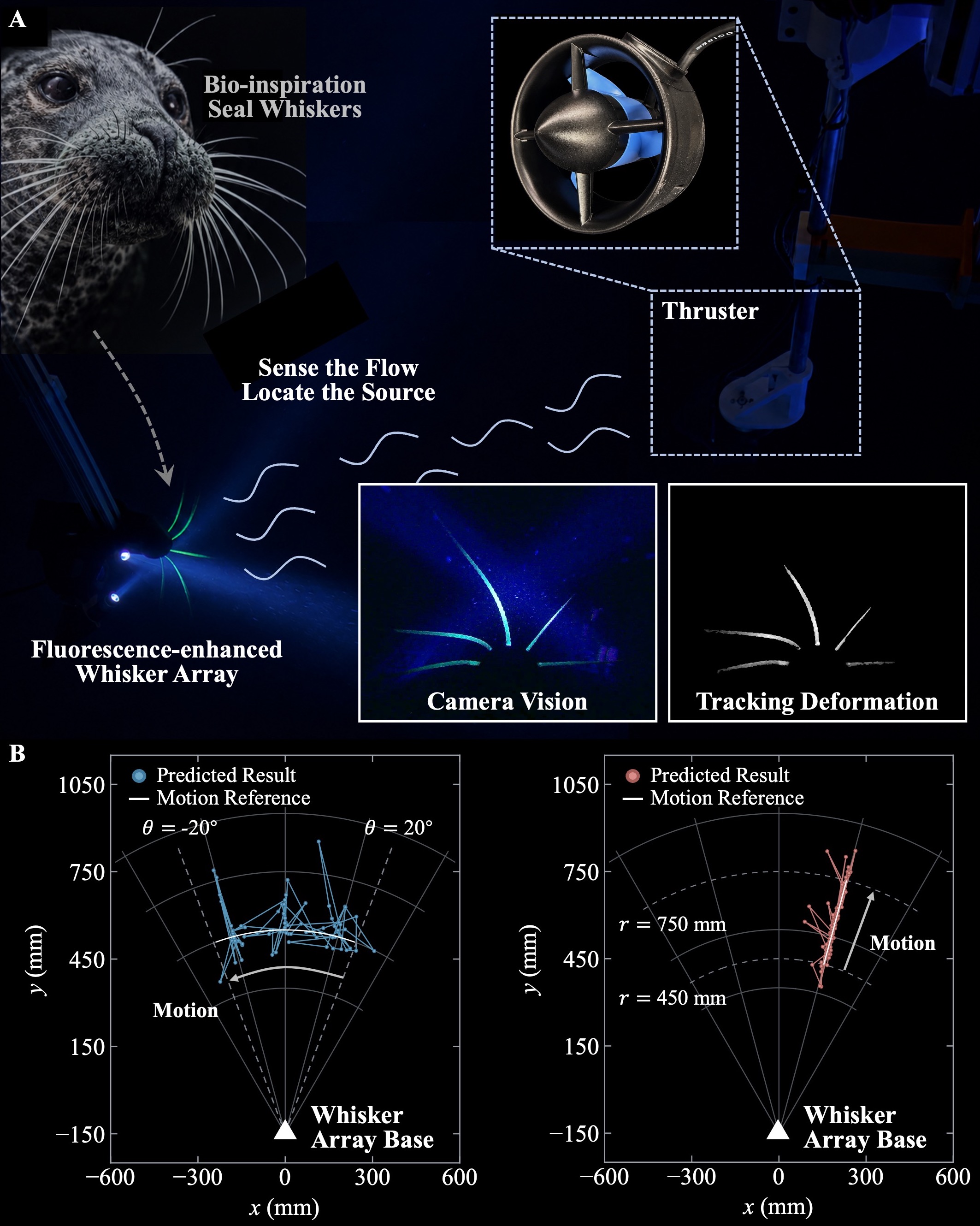}
    \caption{Overview of the proposed fluorescence-enhanced whisker sensing framework. 
    \textbf{(A)} Bio-inspired sensing concept, where flow disturbances generated by an underwater source induce whisker deformation that is captured by monocular vision and used for source localization. 
    \textbf{(B)} Representative moving-source experiments demonstrating real-time localization along different source trajectories. }
	\label{fig::sys_overall}
\end{figure}

Hydrodynamic sensing provides cues by exploiting object-generated disturbances without requiring direct visual observation~\cite{dehnhardt2001trails, yang2006distanttouch}. 
Inspired by the fish lateral line, distributed flow sensors have demonstrated wake detection and source
localization~\cite{qiu2024dipole}. 
Pressure and flow-velocity sensor fusion has enabled dipole-source localization~\cite{jiang2022fusion}. 
Pressure histories have been further used to estimate a moving propeller's lateral state at fixed longitudinal separations~\cite{wang2025propeller}, while distributed
optical flow sensors have reconstructed moving-source
positions~\cite{wolf2019recurrent}. 
However, these distributed methods obtain spatial information using multiple sensors with individual data, which introduces substantial complexity in sensor placement, calibration, synchronization, and system integration, particularly when scaling to larger sensing arrays. 

Inspired by the trail-following capability of harbor seals using their whiskers, researchers have developed compliant sensing mechanisms to perceive hydrodynamic cues and estimate the location of the source~\cite{beem2015slaloming}. 
Several mechanisms are applied to acquire this information based on artificial whiskers. 
Strain-gauge arrays have been utilized to predict dipole-source positions~\cite{geng2025localization}.
Dai et al.~\cite{dai2024magnetic} integrated a three-axis Hall sensor into a magnetic film coupled with a whisker array to estimate flow and wake characteristics, further deducing source direction. 
Glick et al.~\cite{glick2024fbg} developed Fiber Bragg Grating (FBG) sensors to estimate disturbance arrival direction and propagation speed. 
A compliant whisker with root-mounted FBGs has been presented to support wake-guided robotic branch selection~\cite{li2026fbg}. 
However, these approaches typically deploy one dedicated sensor to measure one mechanical response, leading to a bulky system when scaling to larger arrays in order to capture additional features of the source. 

A shared imaging interface offers a solution to retain a collection of responses while reducing the need for multiple sensors~\cite{kent2021whisksight, hu2026vibrissae}.
Zhang et al.~\cite{zhang2025flowsight} proposed FlowSight, utilizing an internal camera to observe the deformation and motion of a tentacle for sensing the speed and direction of flow. 
Wang et al.~\cite{wang2025ciliasight} developed fluorescent silicone cilia with a fisheye camera to estimate local flow velocity and direction. 
These systems infer local flow parameters rather than the coordinates of the source that generates the flow~\cite{zhang2025flowsight,wang2025ciliasight}. 
Elshalakani et al.~\cite{elshalakani2020source} further map the mechanical response of optically tracked whisker tips to the position of an upstream cylinder in a planar setting. 
While static source localization has been demonstrated, real-time localization of dynamically moving sources requires further exploration. 

To this end, we present a fluorescence-enhanced whisker-array sensing system that uses a monocular camera to capture flow-induced whisker deformations and a CNN-based regression model to estimate the position of an underwater source, as illustrated in Fig.~\ref{fig::sys_overall}A. 
A moving-source scenario is presented, achieving real-time localization of a dynamically moving source position, as demonstrated in Fig.~\ref{fig::sys_overall}B. 

The main contributions include: 

(1) a proposed a fluorescence-integrated whisker fabrication process, enabling optical observation under low illumination;

(2) a monocular vision-based platform that tracks the deformation of five spatially oriented whiskers through image enhancement and segmentation;

(3) a lightweight CNN regression framework for static source localization and further demonstration on real-time position sensing of a moving source. 

This paper is organized as follows. 
Section~\ref{sec::design} presents the sensing system design and whisker fabrication. 
Section~\ref{sec::method} presents the vision-based deformation analysis and learning-based source localization pipeline. 
Section~\ref{sec::exp} presents the results of experiments and demonstrates dynamic source localization potential. 
A discussion and conclusions are provided in
Sections~\ref{sec::discuss} and~\ref{sec::conclusion}. 

\section{Design and Fabrication}\label{sec::design}
\subsection{System Overview}\label{subsec::sys}

\begin{figure}[!t]
	\centering
	\includegraphics[width=0.99\columnwidth]{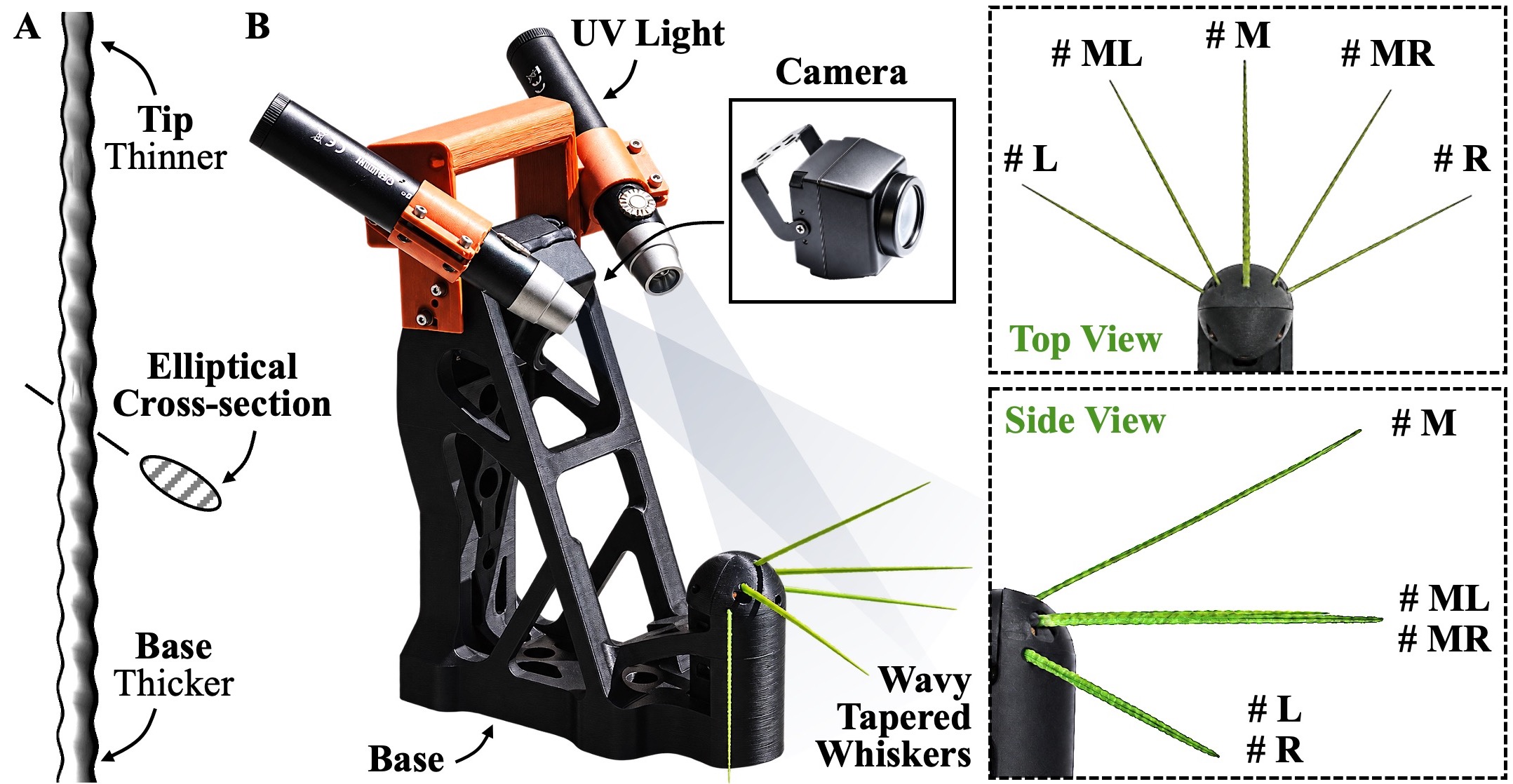}
    \caption{Design and configuration of the fluorescence-enhanced whisker-array sensing system. 
    \textbf{(A)} Bio-inspired compliant whisker with a wavy tapered profile and an elliptical cross-section. 
    \textbf{(B)} Compact sensing platform consisting of a monocular camera, two UV light sources, and an array with five spatially oriented whiskers. 
    The top and side views show the different yaw and pitch orientations of whiskers. }
	\label{fig::design}
\end{figure}

As illustrated in Fig.~\ref{fig::design}A, the key sensing element is a compliant whisker, featuring a wavy geometry, a longitudinally tapered profile, and an elliptical cross-section, which is bio-inspired by seal whiskers. 
This undulating and tapered geometry has been shown to substantially reduce self-excited oscillations as the whisker is dragged through water~\cite{hanke2010vibrissa, beem2015slaloming}. 
The whisker gradually decreases in thickness from the base to the tip. 
These features introduce geometric anisotropy along the whisker body and are utilized to transduce flow-induced hydrodynamic loading into observable deformation. 

Figure~\ref{fig::design}B demonstrates the overall configuration of the proposed fluorescence-enhanced whisker array for underwater source localization. 
The sensing system includes a monocular camera, two fixed low-power ultraviolet (UV) light sources, and an array with five whiskers. 
The camera captures the flow-induced deformation of the whiskers at a resolution of $1920 \times 1080$ pixels at 60 frames/second.
The UV light sources have a nominal $365 \, nm$ excitation and provide a relatively low optical power of approximately $100 \, mW$.
This power is sufficient to produce clearly distinguishable fluorescence from the whiskers for reliable camera-based detection, while limiting illumination and disturbance to the surrounding environment. 
Since the illumination area of a single UV light is small, two sources are arranged on opposite sides of the camera, directed toward the whisker array from different lateral directions to obtain a high-quality view. 
Excluding the soft protruding whiskers, the sensing platform is compact, with overall dimensions of approximately $150 \times 100 \times 250 \, mm$. 

The sensing head contains five whiskers, denoted as $\#L$, $\#ML$, $\#M$, $\#MR$, and $\#R$, respectively, which are arranged with different pitch and yaw angles, as shown in the top and side views in Fig.~\ref{fig::design}B. 
Rather than lying within a common plane, the whiskers span distinct spatial orientations, thereby introducing angular diversity into the array. 
As a consequence, individual whiskers interact differently with the incident flow field according to their orientations relative to the local hydrodynamic loading. 
Different source positions therefore generate distinct combinations of deformation magnitude and direction across the five whiskers. 
Although the monocular camera observes the whisker array from only a single view, and therefore does not directly preserve the complete three-dimensional deformation of each individual whisker, the angular diversity of the whisker array provides complementary deformation information within the image plane. 
Specifically, for a given condition, a deformation component that is weakly observable for one whisker may be pronounced for other whiskers with different pitch and yaw orientations. 
The array responses therefore reduce the directional ambiguity associated with single-view observation and preserve source-dependent deformation patterns that may not be evident from any individual whisker alone. 
This complementary sensing mechanism provides sufficient visual information for subsequent deformation analysis and source localization using a single camera. 

\subsection{Fabrication}\label{subsec::fab}

\begin{figure*}[!t]
	\centering
    \includegraphics[width=0.99\textwidth]{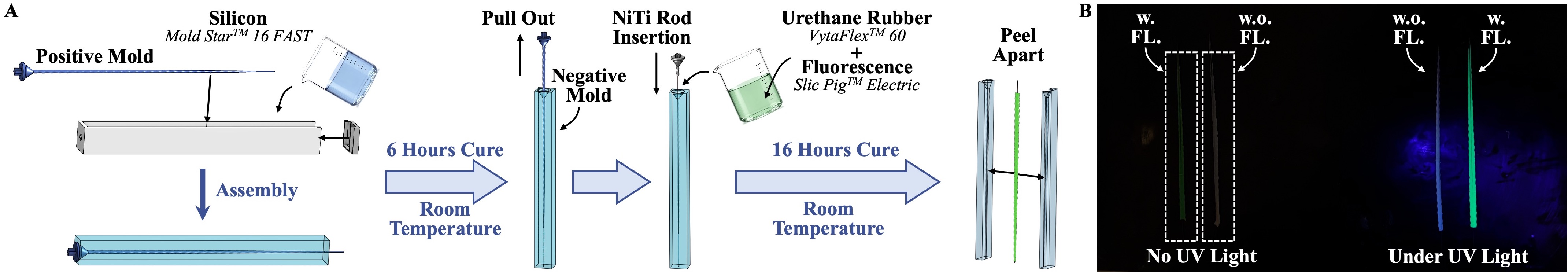}
	\caption{Fabrication process and performance of the fluorescence-enhanced compliant whiskers. 
    \textbf{(A)} Two-stage molding procedure with fluorescence incorporated. 
    \textbf{(B)} Illustrative comparison of whiskers with and without fluorescent pigment under conditions with and without UV illumination. }%
	\label{fig::fab}
\end{figure*}

The fabrication process of the fluorescence-enhanced compliant whisker is depicted in Fig.~\ref{fig::fab}A, which is adapted from the two-stage molding procedures presented in~\cite{li2026fbg}. 
First, a rigid positive mold with designed whisker geometry is 3D-printed and enclosed in a mold box. 
$20 \, g$ silicone rubber ($Mold$ $Star^{TM}$ $16$ $FAST$) is prepared, stirred thoroughly for $3 \, min$ and vacuum-degassed, before being poured around the positive mold. 
The silicone is then cured at room temperature for $6$ hours. 
After curing, the positive mold is withdrawn to obtain a single-use compliant negative mold. 
Subsequently, a superelastic nitinol (NiTi) rod with $0.4 \, mm$ in diameter is inserted into the mold cavity. 
For each whisker, fluorescent pigment is incorporated during the preparation of the urethane overmold. 
A total of $5 - 8 \, g$ casting mixture is prepared, mixed by two parts of urethane rubber ($VytaFlex^{TM}$ $60$) and green fluorescence ($Silc$ $Pig^{TM}$ $Electric$). 
The fluorescent pigment constitutes $5 \%$ by weight of the uncured mixture. 
The pigment is first thoroughly dispersed in \textit{Part A}, then \textit{Part B} is added. 
The mixture is stirred thoroughly for $3 \, min$ and vacuum degassed. 
After the mixture is injected into the silicone mold around the nitinol wire and cured at room temperature for $16$ hours, the silicone mold is peeled apart to release the finished whisker. 
The NiTi rod remains within the fluorescent urethane shell to provide the desired elasticity and toughness.

The effect of incorporating fluorescent pigment is illustrated in Fig.~\ref{fig::fab}B by comparing whiskers with and without fluorescent additives under different illumination conditions. 
Without UV excitation, the visual appearance of the fluorescent whisker remains comparable to that of the non-fluorescent counterpart. 
However, the whisker containing fluorescent pigment exhibits substantially higher visual contrast under UV illumination, making its contour readily distinguishable from the dark background. 
Consequently, a low illumination intensity can be applied to achieve sufficient image contrast for subsequent deformation analysis.
This fluorescence-enhanced contrast enables reliable visual tracking while reducing unnecessary light exposure to the surrounding underwater environment. 

\section{Vision-Based Deformation Analysis}\label{sec::method}

\begin{figure*}[!t]
	\centering
	\includegraphics[width=0.99\textwidth]{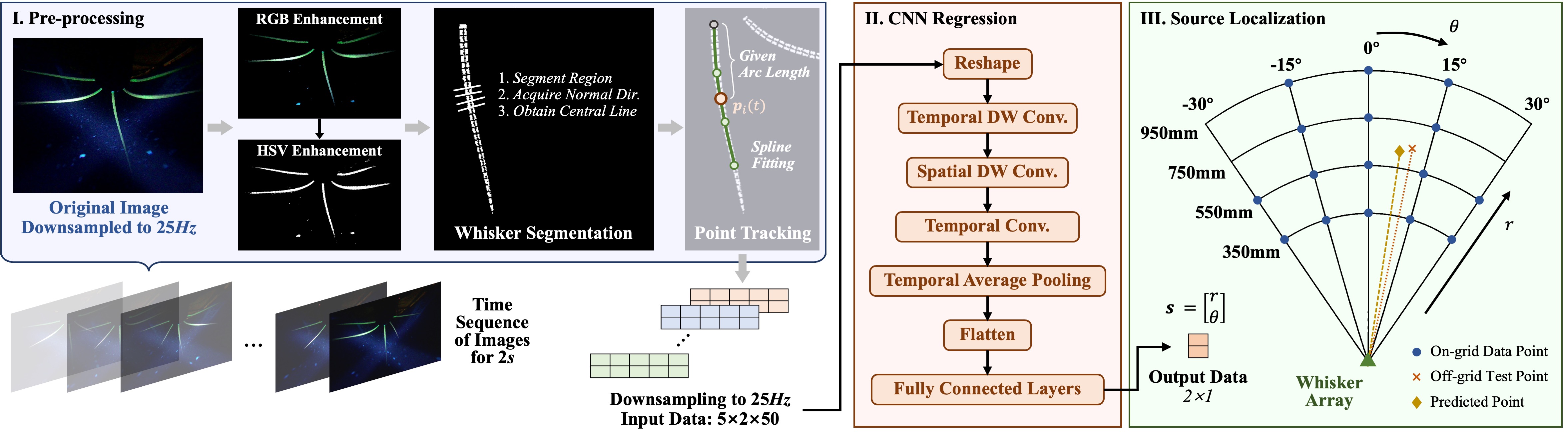}
    \caption{Pipeline of the proposed vision-based deformation analysis and underwater source localization method, including pre-processing, learning-based regression using a lightweight CNN, and source localization. }
	\label{fig::method}
\end{figure*}

The complete pipeline for vision-based whisker deformation analysis and underwater source localization is illustrated in Fig.~\ref{fig::method}. 
The pipeline converts a sequence of whisker images into an array-level deformation representation and subsequently maps the measured deformation to the source position. 
It mainly consists of: image pre-processing, whisker segmentation, intermediate-point deformation tracking, and learning-based regression, enabling efficient and continuous tracking of the whisker array using a single camera and subsequent source localization. 

\subsection{Segmentation and Tracking}\label{subsec::seg}

The deformation analysis is performed on a sequence of images captured by the camera. 
For each frame, image pre-processing is first applied to enhance the fluorescent whiskers and suppress the surrounding underwater background and the UV light. 
Since the embedded fluorescent pigment exhibits a predominantly green response under UV excitation, the $RGB$ channels are combined as $I_{\mathrm{G}} = c_{1} \cdot G - c_{2} \cdot R - c_{3} \cdot B$, where $c_{1} - c_{2} - c_{3} = 1$ and $R$, $G$, and $B$ denote the red, green, and blue channel intensities, respectively. 
This operation enhances pixels dominated by green fluorescence while suppressing the other two colors. 
Subsequently, the saturation and intensity contrast are adjusted to separate the fluorescent whiskers from the relatively dark background. 
The $RGB$ and $HSV$ enhancements are achieved automatically to obtain the best quality by the algorithm. 
Eventually, the whisker regions exhibit substantially higher contrast, providing a more distinguishable representation for the segmentation. 

Based on the enhanced image, the five whisker regions are identified and segmented from the background, providing the pixel sets associated with the individual whiskers. 
For each whisker, several points are extracted along its centerline and used to reconstruct a smooth centerline through spline interpolation. 
Since the whisker root is rigidly attached to the sensing base, its pixel position is assumed to remain fixed during deformation and is therefore used as the reference point for defining the centerline arc length.
Rather than directly tracking the whisker tip, whose large displacement may result in reduced tracking robustness, an intermediate point is selected at a prescribed arc-length position along the reconstructed centerline. 
The pixel coordinate of this tracking point at time $t$ is denoted as $\vect{p}_{i}(t) = [x_{i}(t), y_{i}(t)]^{T}$, where $i \ in \{ L,ML,M,MR,R \}$ denotes the whisker index. 
Before the thruster is activated, the whiskers are recorded under the quiescent condition to determine their baseline positions, denoted as $\vect{p}_{i,0} = [x_{i,0}, y_{i,0}]^{T}$. 
Therefore, the deformation observed by the camera is then represented by the relative pixel displacements as $\Delta \,  \vect{p}_{i}(t) = \vect{p}_{i}(t) - \vect{p}_{i,0} \in \mathbb{R}^{2 \times 1} $. 
Applying this procedure to all five whiskers yields a set of time-dependent data. 
The images, downsampled to a stable $25 \, Hz$, are recorded for $2 \, s$ to obtain sequential input data, which can be expressed in $5 \times 2 \times 50$ dimensions. 
The three dimensions correspond to the five whiskers, two image-plane displacement components, and temporal samples, respectively. 
Each deformation channel is further normalized before being provided to the regression model. 

\subsection{Learning-based Lightweight CNN Regression}\label{subsec::regression}

A lightweight convolutional neural network (CNN) is employed to map the deformation of the whisker array to the source position. 
Firstly, the obtained sequential input data is reshaped into a tensor in $10 \times 1 \times 50$, such that the $x$ and $y$ components of displacements are treated as individual channels. 
After normalization, the first layer performs channel-wise temporal feature extraction using a grouped depth-wise (DW) convolution. 
The convolution consists of $10$ groups, with $8$ independent temporal filters assigned to each channel and a kernel size of $1\times 8$. 
This allows the displacement components of different whiskers to learn independent temporal weights, since their contributions to source localization differ. 
With temporal padding, the sequence length is preserved, and the resulting feature maps are rearranged into $8 \times 10 \times 50$. 
Batch normalization (BN) and an exponential linear unit (ELU) activation are subsequently applied. 

The responses across the whisker array are then fused using a spatial DW convolution with a kernel of $10 \times 1$ and a depth multiplier $4$. 
By spanning all ten deformation channels, this layer combines the independently extracted temporal responses of the five whiskers and produces array-level features in $32 \times 1 \times 50$. 
The output is followed by BN, ELU activation, and dropout. 
A separable temporal convolution is subsequently applied to further extract temporal characteristics from the fused deformation signals. 
It consists of a $1 \times 8$ DW temporal convolution followed by a $1 \times 1$ point-wise (PW) convolution with $64$ filters, yielding the resulting features in $64 \times 1 \times 50$. 
This strategy provides additional temporal feature extraction while substantially limiting the number of trainable parameters compared with a conventional convolution. 
Finally, temporal average pooling with a kernel size of $1 \times 50$ and stride $50$ reduces the temporal dimension to obtain the tensor in $64 \times 1 \times 1$. 
After dropout and flattening, the extracted features are passed through fully connected layers, and the final linear layer directly predicts the source position as $\vect{s} = [r, \theta]^{T}$. 
The combination of grouped, depthwise, and separable convolutions preserves channel-specific and array-level deformation information while maintaining a low model complexity, making the network suitable for source localization with a comparatively small amount of training data. 

\section{Validation and Experiments}\label{sec::exp}

\subsection{Experimental Setup}\label{subsec::setup}

\begin{figure}[!t]
	\centering
	\includegraphics[width=0.99\columnwidth]{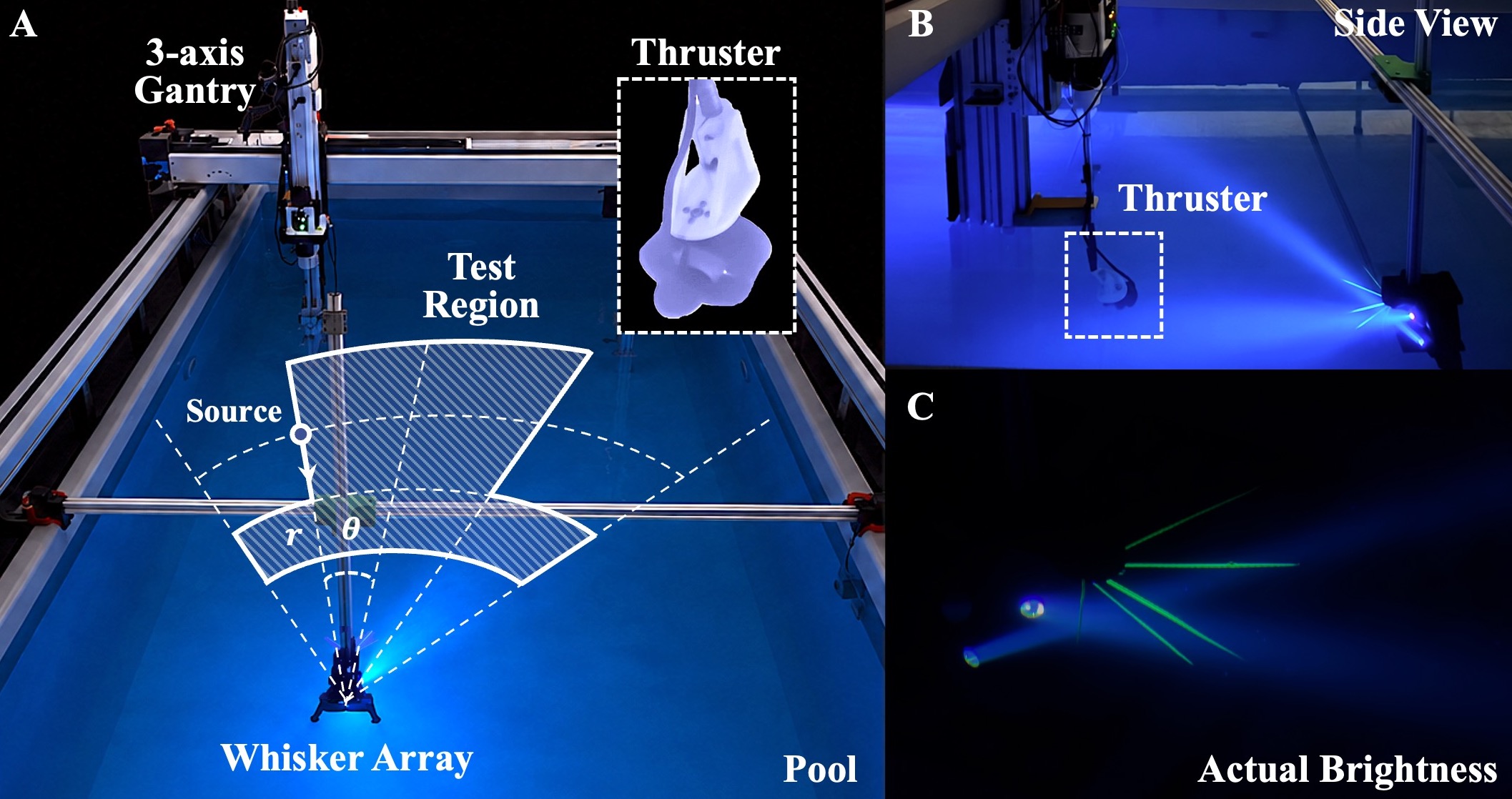}
    \caption{Experimental setup of underwater source localization. 
    \textbf{(A)} Overall configuration, where a thruster mounted on a three-axis gantry is positioned within a fan-shaped test region and continuously oriented toward the whisker array. 
    \textbf{(B)} Side view of the setup. 
    \textbf{(C)} Actual brightness under the illumination condition during data acquisition. 
    }
	\label{fig::setup}
\end{figure}

The experimental setup is shown in Fig.~\ref{fig::setup}. 
A thruster is employed as the underwater source and mounted on a three-axis gantry, which allows its position to be adjusted relative to the whisker array. 
The localization experiments are conducted approximately within the same horizontal plane as the whisker array. 
During all tests, the thruster is oriented toward the whisker array such that the generated flow is directed toward the sensing system, while only the relative source position is varied. 
The thruster is operated under the same input condition throughout all experiments to maintain a consistent source strength across different test locations. 

The source position is parameterized in polar coordinates $\{ r,\theta \}$ with the farthest tip end of the whisker array as the original point. 
Herein, $r$ denotes the radial distance between the thruster and the whisker array, while $\theta$ represents the angular offset of the source relative to the central axis of the sensing system, as illustrated in Fig.~\ref{fig::setup}A. 
The radial distance is varied from $350 \, mm$ to $950 \, mm$ with an interval of $200 \, mm$. 
The angular position ranges from $-30^{\circ}$ to $30^{\circ}$ with an interval of $15^{\circ}$. 
This results in a total of $16$ discrete source locations distributed over the fan-shaped test region, since $4$ locations are beyond the range of the pool. 

Figure~\ref{fig::setup}B provides a side view of the experimental configuration. 
For visualization purposes, the brightness of Fig.~\ref{fig::setup}A and B has been enhanced to make the experimental apparatus and source arrangement clearly visible. 
Figure~\ref{fig::setup}C shows the scene under the actual illumination condition used during data acquisition. 
The surrounding environment remains largely dark, while the fluorescent whiskers are clearly visible under localized UV excitation, providing sufficient image contrast for deformation analysis.

\subsection{Data Collection}\label{subsec::data}

For each source location, the thruster is first moved to the prescribed position by the three-axis gantry and reoriented to face the whisker array. 
After positioning, the thruster remains inactive for $10 \, s$ to allow the surrounding water and whiskers to return to a quiescent state. 
The camera then records the initial positions of the five whiskers as the reference configuration for subsequent deformation analysis. 
The thruster is then activated using a ramp-up profile rather than a step input, thereby reducing strong transient disturbances during startup. 
After reaching the prescribed operating intensity, the thruster is maintained at the same input condition for an additional $10 \, s$ to allow the induced flow field and whisker responses to stabilize. 
The camera then records the whisker deformation continuously for $90 \, s$. 
After recording, the thruster is ramped down and moved to the next source location, where the same acquisition procedure is repeated. 
Each recording is divided into $45$ non-overlapping $2 \, s$ segments. 
Each $2$-s segment is treated as one localization sample. 
One complete round of test consists of measurements at all $16$ locations, and the entire procedure is repeated for $12$ rounds in two days to construct the data set. 
Therefore, each source location provides $45 \times 12 = 540$ training samples, resulting in a total of $8640$ samples. 
It should be noted that a five-fold cross-validation strategy is employed for model training and validation. 

\begin{figure}[!t]
	\centering
    \includegraphics[width=0.99\columnwidth]{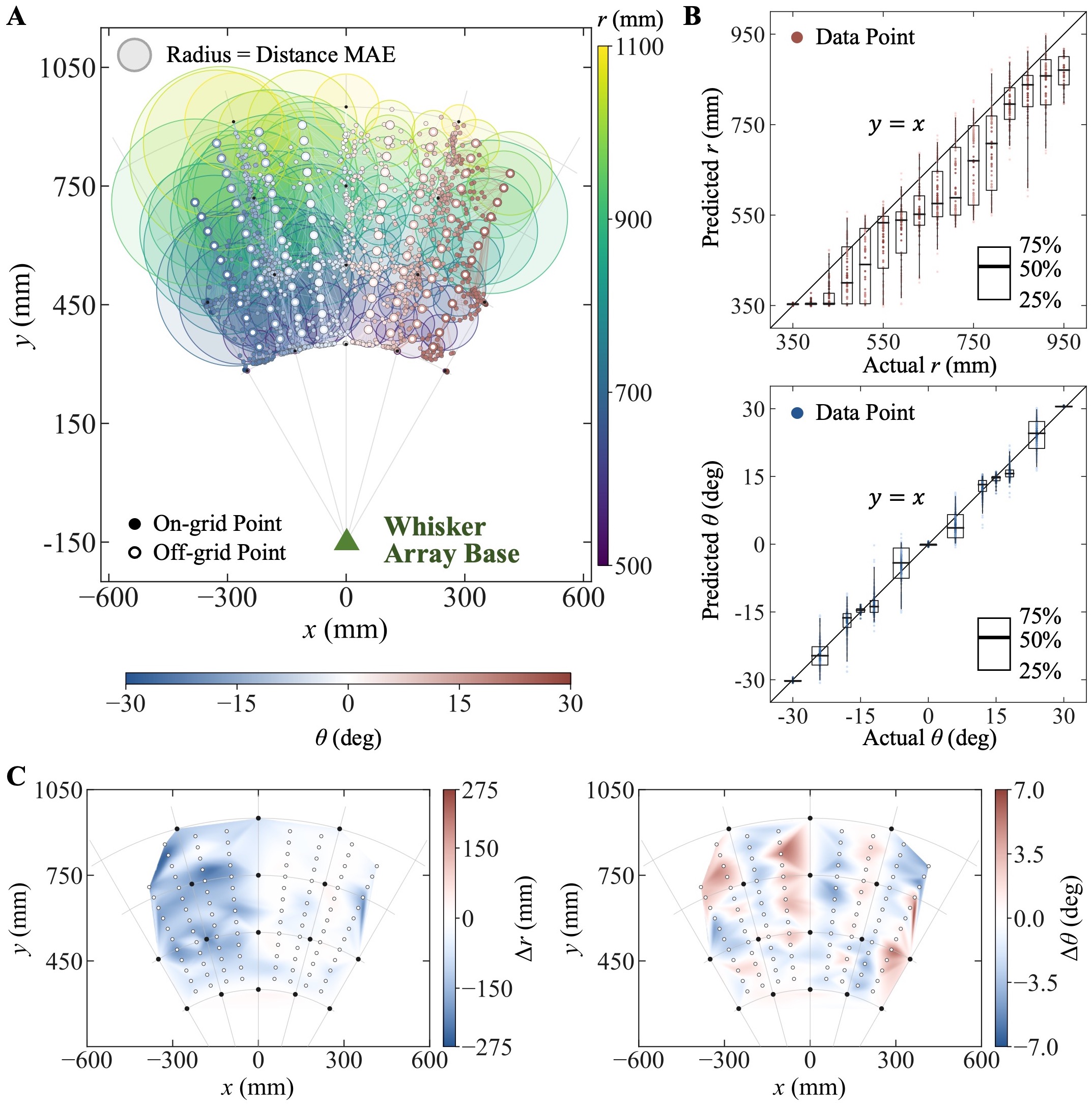}
	\caption{Source localization performance. 
    \textbf{(A)} Spatial distribution of the predicted source locations. 
    Filled and hollow markers indicate the on-grid and off-grid locations. 
    The radius of each error circle denotes the mean absolute localization error at the corresponding location.
    \textbf{(B)} Predicted versus actual distance $r$ and angle $\theta$. 
    The diagonal line indicates the ideal prediction $y=x$, and the box plots summarize the $25 \, \%$, $50 \, \%$, and $75 \, \%$ percentiles of the predictions at each test coordinate. 
    \textbf{(C)} Spatial distributions of the radial error $\Delta r$ and angular error $\Delta \theta$ over the test region. }
	\label{fig::verify}
\end{figure}

To evaluate localization performance beyond the discrete grid, an independent test set is collected at intermediate source locations, including all $16$ on-grid and $108$ off-grid points. 
Specifically, the distances range from $350 \, mm$ to $950 \, mm$ with an interval of $40 \, mm$ while the angles range from $-30^{\circ}$ to $30^{\circ}$ with an interval of $6^{\circ}$. 
The same acquisition protocol is applied with $10$ rounds of $2 \, s$ data points are recorded for each location. 
It is worth noting that the evaluation not only examines the performance on the same discrete locations with measured data, but also tests whether the continuous mapping can interpolate deformation patterns associated with previously unseen source positions. 

\subsection{Source Localization Performance} \label{subsec::verify}

The localization performance on the test set is summarized in Fig.~\ref{fig::verify}. 
Figure~\ref{fig::verify}A presents the predicted source locations in the test region plane. 
The ground-truth on-grid locations are indicated by small filled black markers, whereas the off-grid locations are represented by hollow markers. 
For each source location, the ten predictions are shown as filled data points. 
To visualize the location-dependent prediction accuracy, a circle is centered at each actual location with its radius determined by the Mean Absolute spatial localization Error (MAE) between the predicted and actual positions. 
The on-grid cases exhibit smaller errors than the off-grid ones, which is expected since the off-grid predictions are the result of interpolation through the trained model. 
The relatively concentrated off-grid predictions indicate that the learned deformation-to-position mapping is not restricted to discrete training locations and retains the ability to estimate intermediate source positions. 
Furthermore, a mild spatial asymmetry can be observed, with generally smaller errors on the right side of the test region than on the left. 
One possible explanation is the asymmetric placement of the whisker array within the pool, where the left side is closer to the wall such that boundary-induced flow reflections and disturbances may perturb the local flow field and reduce the consistency of the source-dependent deformation patterns, leading to larger localization errors.

\begin{table}[!t]
    \centering
    \caption{Quantitative localization errors on the testing set.}
    \label{tab::performance}
    \resizebox{\columnwidth}{!}{
    \begin{tabular}{cccc}
        \toprule
        & $\Delta r_{MAE}$ (mm) 
        & $\Delta \theta_{MAE}$ ($^\circ$) 
        & $\Delta d_{MAE}$ (mm) \\
        & (Avg/Min/Max) & (Avg/Min/Max) & (Avg/Min/Max) \\
        \midrule
        On-grid 
        & 45.2 / 0.1 / 202.6
        & 0.47 / 0.02 / 3.76 
        & 47.2 / 0.9 / 202.8 \\
        Off-grid 
        & 77.7 / 0.1 / 278.6
        & 2.78 / 0.01 / 11.75
        & 94.2 / 5.5 / 279.2 \\
        \midrule
        All (\textbf{Avg})
        & \textbf{73.5}
        & \textbf{2.48}
        & \textbf{88.1} \\
        \bottomrule
    \end{tabular}}
\end{table}

Figure~\ref{fig::verify}B further examines the prediction results by comparing the estimated and actual values of the radial distance $r$ and angle $\theta$. 
For each value of $r$ (or $\theta$), all test samples sharing the same $r$ (or $\theta$) are grouped together regardless of their corresponding $\theta$ (or $r$), so that the overall prediction distribution and trend along each coordinate can be evaluated independently. 
The solid diagonal line $y=x$ indicates an ideal result. 
The lightly colored data points show the individual predictions, and the box plots summarize their distributions at each location using the $25 \, \%$, $50 \, \%$, and $75 \, \%$ percentiles, with the central line denoting the median. 
For both $r$ and $\theta$, the prediction medians generally follow the identity line. 
A noticeable tendency is that the angular estimates exhibit a relatively narrow prediction distribution at all test locations, whereas the range estimates show substantially larger dispersion at off-grid locations. 
At these intermediate positions, the predicted $r$ is biased toward the neighboring on-grid $r$. 
This suggests that the proposed model learns the angle-dependent deformation features more effectively, resulting in a more accurate mapping from the whisker response to the angular location of the source. 
In contrast, the radial deformation cues are less distinctive, leading to a weaker mapping and a more pronounced interpolation bias. 
Detailed quantitative error metrics are summarized in Table~\ref{tab::performance}, where $d$ denotes the distance between the predicted and actual locations. 
Despite these differences, the model retains a promising interpolation capability, allowing source locations to be estimated continuously across the test region. 

Figure~\ref{fig::verify}C further visualizes the spatial distribution. 
For the angular error, alternating regions of positive and negative bias can be observed between neighboring on-grid locations. 
Meanwhile, the radial error exhibits a more pronounced spatial variation, consistent with the tendency in Fig.~\ref{fig::verify}B, where predictions at off-grid locations are generally more negative. 
This behavior may partly be attributed to the training objective, which intentionally assigns greater importance to angular accuracy than to radial accuracy, reflecting the design preference for reliable source direction over precise range estimation in the localization task. 

\subsection{Demonstration of Moving-Source Localization} \label{subsec::demo}

\begin{figure*}[!t]
	\centering
    \includegraphics[width=0.99\textwidth]{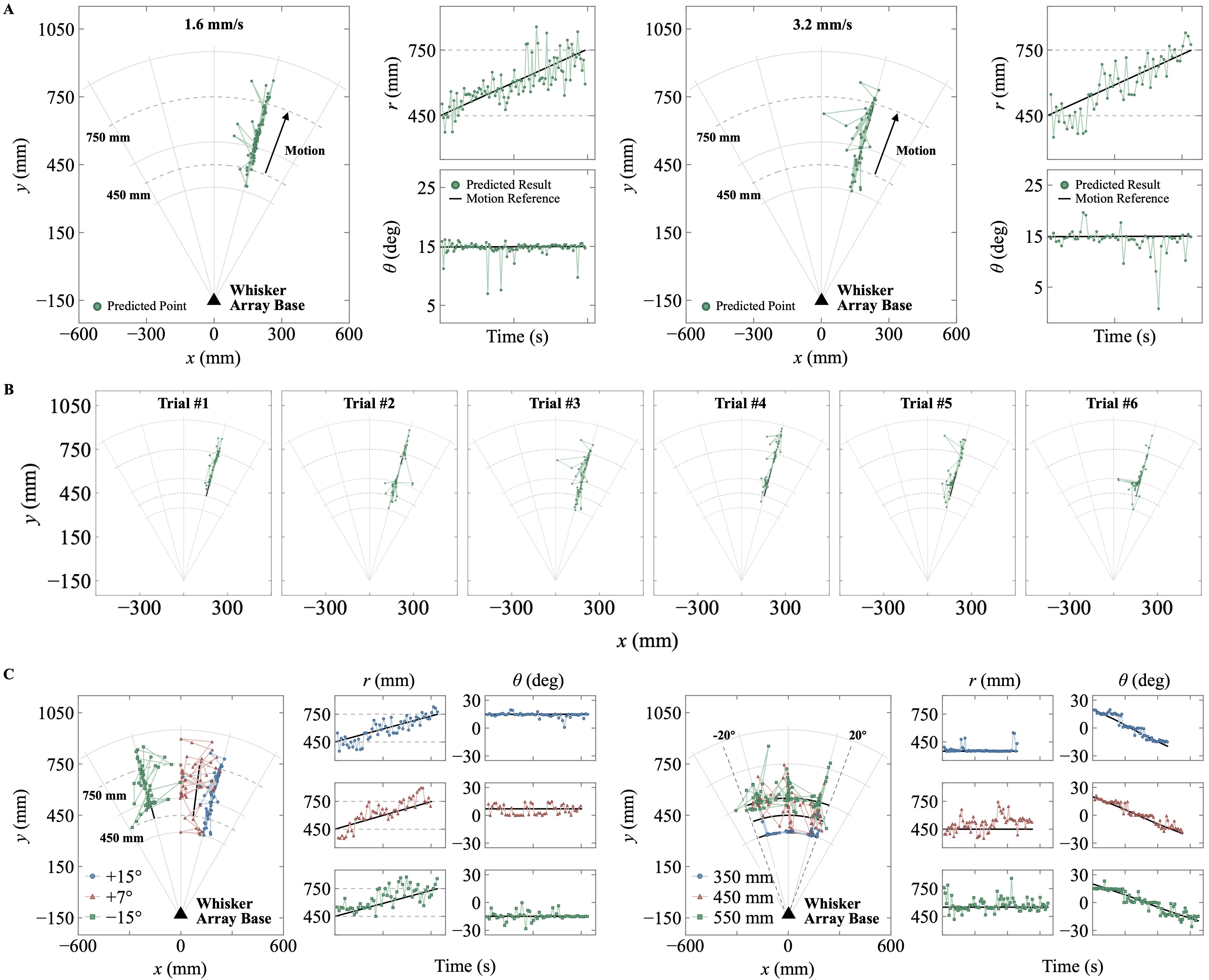}
	\caption{Demonstrations and performance of a moving-source dynamic localization. 
    \textbf{(A)} Comparison under different motion speeds, from quasi-static to more dynamic conditions. 
    \textbf{(B)} Repeated trials demonstrating repeatability performance of localization. 
    \textbf{(C)} Representative real-time localization of different source trajectories, together with the estimated radius $r$ and angle $\theta$. }
	\label{fig::demo}
\end{figure*}

To further evaluate the applicability of the trained model to dynamic scenarios, moving-source experiments are conducted under different motion trajectories and speeds, as shown in Fig.~\ref{fig::demo}. 
First, the source moves along the same radial path from $450 \, mm$ to $750 \, mm$ at an on-grid angle $15^{\circ}$ with a speed of $1.6 \, mm/s$, corresponding to a displacement of $3.2 \, mm$ within each $2 \, s$ inference window and therefore representing a near-quasi-static condition. 
As shown in Fig.~\ref{fig::demo}A, the predictions perform well with $\Delta r_{MAE} \approx 51.7 \, mm$ and $\Delta \theta_{MAE} \approx 0.63^{\circ}$.
Furthermore, the same trajectory is repeated at a higher speed of $3.2 \, mm/s$ for comparison. 
Larger deviations are observed at the higher speed, which may be associated with stronger motion-induced variations in the local flow characteristics and whisker responses. 
Figure~\ref{fig::demo}B depicts six repeated runs along the same trajectory. 
The results exhibit similar prediction trends, demonstrating consistent dynamic prediction performance. 
The largest $\Delta r_{MAE}$ among the six trials is $85.0 \, mm$ and the largest $\Delta \theta_{MAE}$ is $1.65^{\circ}$. 

The moving-source experiments are further extended to trajectories with different spatial parameters. 
For radial motion, two on-grid angles ($-15^{\circ}$ and $15^{\circ}$) and one off-grid angle ($7^{\circ}$) are tested. 
Although the model follows the radial motion in all three cases, larger prediction deviations are observed for the off-grid trajectory, consistent with the interpolation errors observed in the static tests. 
The largest $\Delta d_{MAE}$ is $105.3 \, mm$ and the largest $\Delta \theta_{MAE}$ is $4.91^{\circ}$. 
The smaller MAE than in the static tests is mainly due to the more limited radial distance of the trajectory, which excludes the edge regions associated with larger localization errors.
For arc-shaped motion, trajectories at different fixed radial distances are tested, with two on-grid radii ($350 \, mm$ and $550 \, mm$) and one off-grid radius ($450 \, mm$). 
Similarly, larger deviations are observed for the off-grid trajectory. 
The largest $\Delta d_{MAE}$ is $94.3 \, mm$ and the largest $\Delta \theta_{MAE}$ is $3.03^{\circ}$. 
The relatively smaller MAE is also due to the restricted angular range of the trajectories, excluding the boundary regions. 

Overall, although the localization model is trained solely using stationary-source data, it continuously provides effective source-position estimations under relatively dynamic scenarios. 
These results demonstrate the potential of the proposed framework for real-time localization of moving hydrodynamic sources, while the increased errors under faster motion also indicate the need for dedicated dynamic training data for more accurate continuous localization.

\section{Discussion}\label{sec::discuss}

\textit{Limitations: }
The proposed source localization approach is limited to planar localization, with the thruster approximately coplanar with the whisker array, oriented toward the whisker array, and operated at a fixed power level. 
Extending the framework to more general scenarios would require additional source variables, such as height, heading, and strength, together with expanded training data and an updated model. 
Increasing the density and coverage of on-grid calibration locations could further improve off-grid interpolation. 
Moreover, the model is primarily trained on stationary-source data under quasi-steady conditions. 
Dynamic localization and continuous tracking would therefore benefit from dedicated trajectory-based training data. 

\begin{figure}[!t]
	\centering
    \includegraphics[width=0.99\columnwidth]{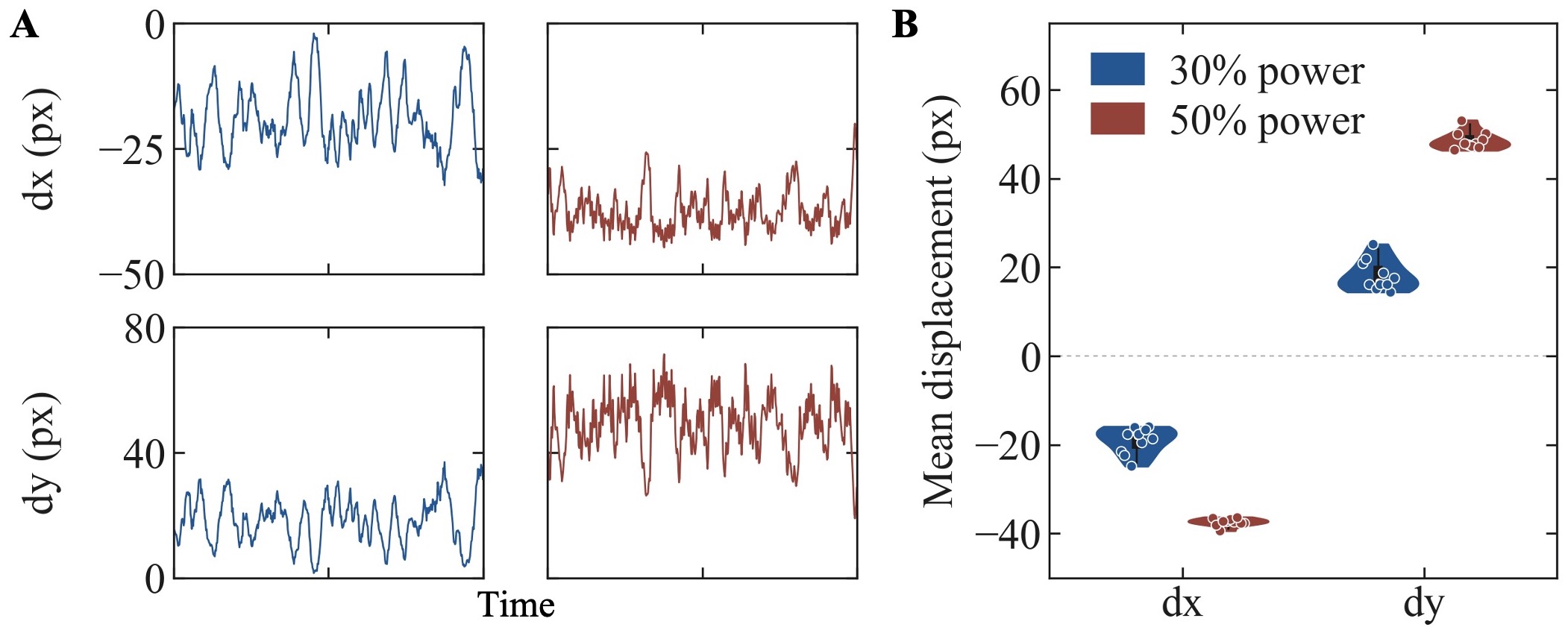}
	\caption{Comparisons of whisker deformation under thruster with different power. 
    \textbf{(A)} Original data sequence. 
    \textbf{(B)} Average displacement. }
	\label{fig::compare}
\end{figure}

To provide preliminary insight into the influence of source intensity, the thruster is operated at a higher power level at an on-grid location with $r = 750 \, mm$ and $\theta = 15^{\circ}$. 
A $20 \, s$ sequence is recorded, and the displacement of the tracking point on one whisker is compared with that under the smaller power. 
Figure~\ref{fig::compare}A shows the displacement time series, while Fig.~\ref{fig::compare}B compares the mean deformation calculated over consecutive $2 \, s$ segments. 
The higher-power excitation produces a larger mean whisker deformation with reduced variation, suggesting a stronger and more consistent mechanical response that may facilitate more reliable sensing. 

\textit{Future Work: }
To further visualize and interpret the hydrodynamic mechanisms underlying the measured whisker responses, particle image velocimetry (PIV) could be employed to characterize the wake and local flow structures around the whisker array. 
Such measurements would provide additional insight into the relationship among source position, surrounding flow features, and the resulting whisker deformation patterns. 
The localization framework could also be further improved. 
Additional depth cues could be incorporated and a higher frame rate would provide richer temporal and frequency-domain characteristics of the whisker response. 
The source set could also be expanded to include different types and operating characteristics, such as varying strengths and excitation frequencies, enabling the system to distinguish and localize a broader range of underwater hydrodynamic sources in real time. 
More broadly, the proposed vision-based whisker-array sensing framework could be integrated into UUVs as a compact hydrodynamic perception module, with the potential to support real-time detection, localization, and continuous tracking of moving sources, especially organisms, through the flow disturbances they generate. 

\section{Conclusion}\label{sec::conclusion}

This paper presents a fluorescence-enhanced whisker array for underwater source localization in low-light environments.
A monocular camera reads flow-induced whisker deformations under local UV excitation, and a CNN-based prediction model estimates source location from whisker deformation image sequences. 
Pool experiments achieve distance accuracy of $88 \, mm$ with $\Delta r_{MAE} \approx 73.5 \, mm$ and $\Delta \theta_{MAE} \approx 2.5^{\circ}$ across a test region of $600 \, mm \times$ with $\pm 30^{\circ}$. 
Moving-thruster demonstrations further showcase the potential capacity of locating source positions in real-time applications. 
These results further highlight the potential of integrating the proposed sensing framework into underwater mobile robots as a compact perception module for continuous detection and tracking of moving hydrodynamic sources. 

\bibliographystyle{IEEEtran}
\normalem
\bibliography{Ref}

\vspace{11pt}

\vfill

\end{sloppypar}
\end{document}